\documentclass[10pt,conference,a4paper]{IEEEtran}

\IEEEoverridecommandlockouts
\usepackage{cite}
\usepackage{amsmath,amssymb,amsfonts}
\usepackage{algorithmic}
\usepackage{graphicx}
\usepackage{textcomp}
\usepackage{xcolor}
\usepackage[T1]{fontenc}

\usepackage{graphicx} 
\usepackage{caption,subcaption}

\usepackage{dblfloatfix}

\def\BibTeX{{\rm B\kern-.05em{\sc i\kern-.025em b}\kern-.08em
    T\kern-.1667em\lower.7ex\hbox{E}\kern-.125emX}}
\usepackage{float}

\begin{document}

\title{ Multi-Stage CNN-Based Monocular 3D Vehicle Localization and Orientation Estimation

}


\author{
\IEEEauthorblockN{Ali Babolhavaeji and Mohammad Fanaei}
\IEEEauthorblockA{Department of Electrical and Computer Engineering and Computer Science \\
University of Detroit Mercy, MI, U.S.A.\\
Email: \{babolhal, fanaeimo\}@udmercy.edu}
}

\maketitle

\begin{abstract}
This paper aims to design a 3D object detection model from 2D images taken by monocular cameras by combining the estimated bird's-eye view elevation map and the deep representation of object features. The proposed model has a pre-trained ResNet-50 network as its backend network and three more branches. The model first builds a bird's-eye view elevation map to estimate the depth of the object in the scene and by using that estimates the object's 3D bounding boxes. We have trained and evaluate it on two major datasets: a syntactic dataset and the KIITI dataset.
\end{abstract}

\begin{IEEEkeywords}
Monocular cameras, 3D object detection, 3D pose estimation, bird's-eye view, computer vision, convolutional neural networks, synthetic dataset, KITTI dataset, TransferLearning.
\end{IEEEkeywords}

\section{Introduction}
Visual object perception is one of the fundamental challenges in computer vision, especially in autonomous driving and robotics. Detecting objects in the 3D space is instrumental in planing and executing a safe route for motion planning algorithms, where depth information is a critical parameter. A large number of self-driving systems exploit different data sources to compensate for the lack of depth information by using expensive range sensors such as LIDAR~\cite{Li2019,Li2016,chen2017multi}. Such systems operate based on the analysis of a point cloud, which is computationally expensive.
Recent research studies have attempted to replace these costly sensors with simpler ones like monocular cameras \cite{liu2019deep ,qin2019monogrnet}, which are widely deployed on most vehicles for other purposes. In comparison, visual perception based on monocular images is noticeably more difficult due to the lack of depth information. Deep neural networks have proven to be highly capable in performing visual perception tasks using such cheap sensors in recent years.\cite{ bertoni2019monoloco ,zhong2019robust}

Over the last few years, we have seen that convolutional neural networks (CNN) have made enormous progress in reliable 2D object detection. The KITTI benchmarks~\cite{geiger2013vision} cite reports that state-of-the-art algorithms can achieve average precision~(AP) of around~$90\%$~\cite{geiger2012we}. Most of the recent works are based on Faster R-CNN because its region proposal network (RPN) efficiently generates object proposals~\cite{ ren2015faster, lin2017feature,Yu2016UnitBox, Li2019, ghalehnovi2014integration, Manhardt2019roi ,newaz2018network , Chen2018}. The RPN is a sliding window that checks all possible spatial locations on an image and finds specific predefined shape templates, known as anchors. Using state-of-the-art methods in the relatively mature field of 2D object detection algorithms, 2D bounding boxes around the detected objects can be estimated with high accuracy, which can significantly reduce the computational expenses of extracting 3D properties.
The objective of this paper is to develop an efﬁcient CNN-based framework to obtain the properties of the 3D bounding boxes of the underlying objects (including the 3D location, dimensions, and orientation) based on the corresponding frontal 2D bounding boxes. The depth of each object is estimated at the intermediate layers so that a bird's-eye view map, i.e., a top view spatial occupancy map of the environment, can be generated by training the network to find the transformation between the frontal 2D bounding boxes and corresponding top-view 2D bounding boxes. Our ultimate goal, in summary, is to determine the 3D bounding boxes of the underlying objects from a monocular image without using point clouds or stereo data.

It is almost impossible to collect a dataset in real-world that has a unified front view and top view of a wide range of scenes. That is why we based our analysis on a synthetic dataset, which is called surrounding vehicles autonomous (SAV) dataset~\cite{Palazzi2017}. To collect this dataset, the researchers have used the environment of the Grand Theft Auto V (GTAV) video game along with an automatic script that toggles between frontal and bird's-eye views at each time step, filtering the inaccurate data in the process. Pretrained network on the SAV dataset is finetuned on the KITTI dataset to create a bird's-eye view grid map. The KITTI  dataset has 7481 training images, which are split in half for for training and validation.

The rest of this paper is organized as follows: Section~\ref{Sec:RelatedWorks} summarizes the main results in the literature on 3D object detection. In Section~\ref{Sec:Model}, we introduce the model of the proposed network and discuss its various branches. The results of the performance evaluations are presented in Section~\ref{Sec:Results}, and Section~\ref{Sec:Conclusions} concludes the paper.

\section{Related works}
\label{Sec:RelatedWorks}
The main 3D object detection algorithms can be categorized based on the input data type as follow: 
\begin{itemize}
   \item  Methods based on the point cloud. typically, project point cloud to the bird's-eye view or front view and use a 2D CNN to extract its features or they will build voxel grids and train a 3D CNN.

   \item  Approaches based on multi-view images can obtain a depth map by calculating the difference among other views.
   \item  Algorithms based on monocular images use a single front-view image to estimate objects locations, which make them the most challenging variant due to the lack of depth information in original 2D images.
\end{itemize}
The multi-view 3D (MV3D) network proposed in \cite{ chen2017multi} is a sensory-fusion framework that uses both point cloud and RGB images at the same time to extract 3D candidate boxes from the bird's-eye view representation of the point cloud. A fully convolutional neural network is proposed in~\cite {Li2016} to project the point cloud to the front view image and build a 2D point map. The algorithm proposed by Chen et~al.~\cite{Chen2017}, known as the 3D object proposal (3DOP), employs a CNN-based model to predict object properties based on stereo images with additional scene priors. PoseCNN proposed in~\cite {xiang2017posecnn} localizes the object's center and depth from the camera for estimating 3D object transformation.
Naiden et~al.~\cite{naiden2019shift} apply the Faster R-CNN network to regress 2D and 3D properties using a geometrically constrained deep learning approach. In the approach suggested by Manhardt et~al.~\cite{Manhardt2018ECCV}, 4D quaternion is found by regression to depict the 3D rotation of an object. A differentiable render-and-compare loss is introduced in~\cite{kundu20183d} to form 3D-RCNN network in order to estimate the full 3D shape of cars and their orientation in the KITTI dataset. The single-shot detector (SSD) framework is extended into SSD-6D in~\cite{ Kehl2017ssd6d } to estimate a 6D hypothesis from RGB images.
We generally follow the semantic-aware dense projection network (SDPN) framework proposed in~\cite{Palazzi2017}, which has focused on a CNN-based model to transform front view images to bird's-eye view perspectives by building a huge synthetic dataset. We use SDPN  to provide depth information as an auxiliary branch and add a new regression branch at the top of SDPN to estimate the 3D properties of each object. We show that by combing the estimated bird-eye-view elevation map and the deep representation of object features, we can  estimate 3D object bounding boxes.

\begin{figure*}[tbh]
    \centering
    \includegraphics[width=1\textwidth]{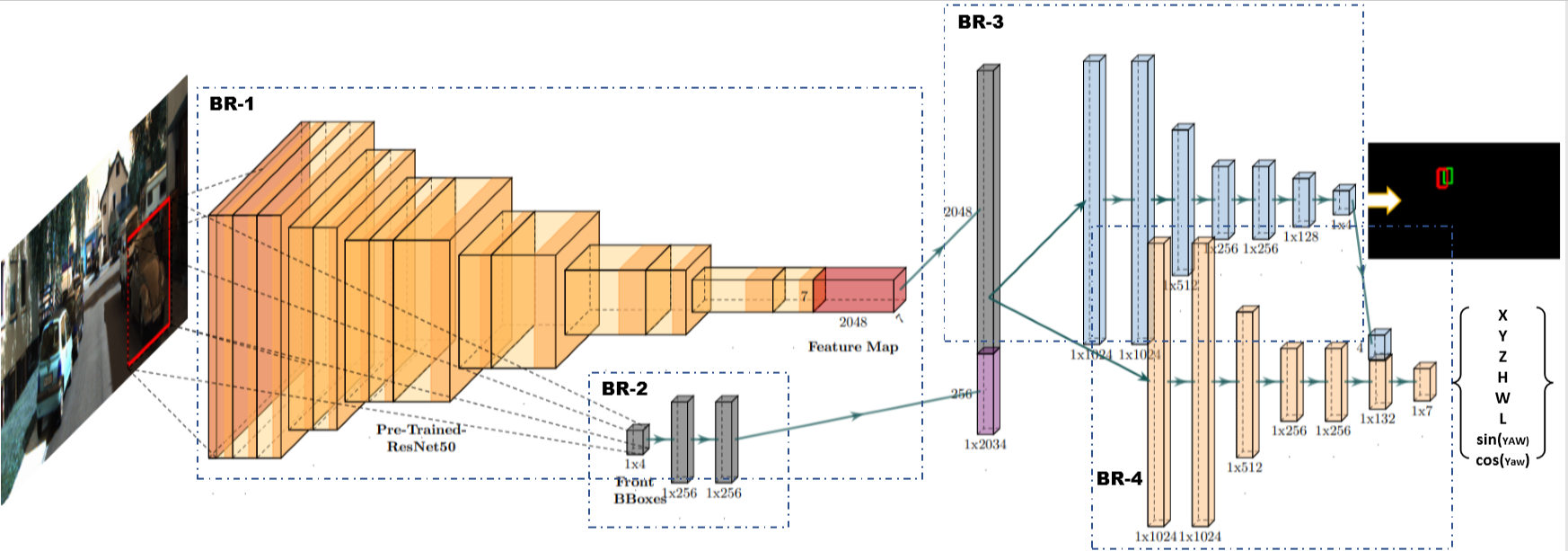}
    \caption{
    The architecture of the proposed model: Branch 1, denoted by ``BR-1'', is shown at the top-left and is the backend of our network. It is based on ResNet-50 with the removal of the last FC layers to produce a deep representation of the object instance (e.g., the red rectangle area in the left-side input image). Branch 1 has $2048\times7\times7$, and the global feature pooling converts it to 2048 feature vectors. Branch 2, denoted by ``BR-2'', is a stack of four FC layers to encode 4-D front bounding boxes to 256-D. Branch 3, denoted by ``BR-3'' and shown on top-right, is an MLP with eight FC layers. At the output of BR-3, the bird's-eye view gird map is created by a post-processing function. Branch 4, denoted by ``BR-4'', is a regression network to predict the properties of the 3D bounding box of the objects, including $p = [X, Y, Z]$ defined as the center point location, $s = [W, L, H]$ defined as the width, length, and height, as well as the orientation of the 3D bounding box, denoted by $\mathsf{Yaw}$.
    }
    \label{fig:Our-model}
\end{figure*}

\section{Our Model}
\label{Sec:Model}
\subsection{Architecture}
Our model is composed of four main branches, depicted in Figure~\ref{fig:Our-model}. In the first branch, there is a backend network that generates a downscaled feature map from the front-view cropped image. Initializing a network by a pre-trained network has shown to improve prediction accuracy even if the pre-trained model and the trained data are hugely different~\cite {gupta2014learning}. We have adopted a pre-trained ResNet-50 network~\cite{he2016deep} trained on ImageNet~\cite{deng2009imagenet} as the backend network. The fully-connected layer of the backend network is discarded to achieve a deep representation of the object instance.
The second branch of the proposed network is an auxiliary multi-layer perceptron~(MLP) to map the corner points of the front view 2D object bounding box (i.e., top-left and bottom-right) to a 256-dimensional space. It contains four fully-connected (FC) layers and is denoted by ``BR-2'' in Figure \ref{fig:Our-model}.
The third branch of the proposed architecture is also an MLP with eight FC layers. It accepts as its input the semantic feature vector generated by concatenating the flatten output of the backend network and the 256-encoded output of BR-2. The outputs of BR-3 are the top-left and bottom-right corners of the bounding boxes in the top-view interpretation of the frontal view for the given object. The output layer in this branch uses ``$\tanh(\cdot)$'' as its activation function, which bounds the outputs to $[-1,1]$. The bird's-eye view elevation grid map will be generated based on the output of BR-3 in the post-processing function, similar to the approach suggested in the SDPN framework~\cite{Palazzi2017}.

The last branch, denoted by ``BR-4'' in Figure~\ref{fig:Our-model}, is the \textit{regression branch}, which is responsible for estimating the properties of the objects’ 3D bounding box, including $p = [X, Y, Z]^T$ defined as the center point location, $p = [W, L, H]^T$ defined as the width, length, and height, as well as the orientation of the 3D bounding box, denoted by $\mathsf{Yaw}$. All of these values are relative with respect to the camera coordinate system. Similar to BR-3, this branch is also formed by eight FC layers but its last FC layer has seven outputs. It regresses objects’ 3D proprieties from the semantic feature vectors.

High-level features related to the 3D scene understanding drastically depend on the depth information~\cite {brazil2019m3d}. We will use the predicted data in BR-3 to calculate the object's depth and use it in the penalty term to train model as a depth-aware model. To this end, we need to have a two-stage training procedure. At the first stage, the model from BR-1 to BR-3 is trained. The trained model at this stage can create the bird's-eye view elevation map at the output of BR-3, similar to SDPN. The bird's-eye view map provides depth information that is used as an additional term in the loss function for training BR-4. At this stage, the weights of the entire model freeze except for BR-4, and the model is trained with a new loss function to achieve a depth-aware architecture.

\subsection{Dataset Analysis}
In this section, we briefly analyze the datasets that are used in this paper.

It is almost impossible to collect a dataset in real-world that has a unified front view and top view of a wide range of scenes. That is why we based our analysis on a synthetic dataset, which is called surrounding vehicles autonomous (SAV) dataset~\cite{Palazzi2017}. To collect this dataset, the researchers have used the environment of the Grand Theft Auto V (GTAV) video game along with an automatic script that toggles between frontal and bird's-eye views at each time step, filtering the inaccurate data in the process. Pretrained network on the SAV dataset is finetuned on the KITTI dataset to create a bird's-eye view grid map. The KITTI  dataset~\cite{geiger2012we} has around 7500 training images, which are split in half for for training and validation.

\begin{figure}[th]
\centering
\begin{subfigure}[b]{.49\linewidth}
\includegraphics[width=\linewidth]{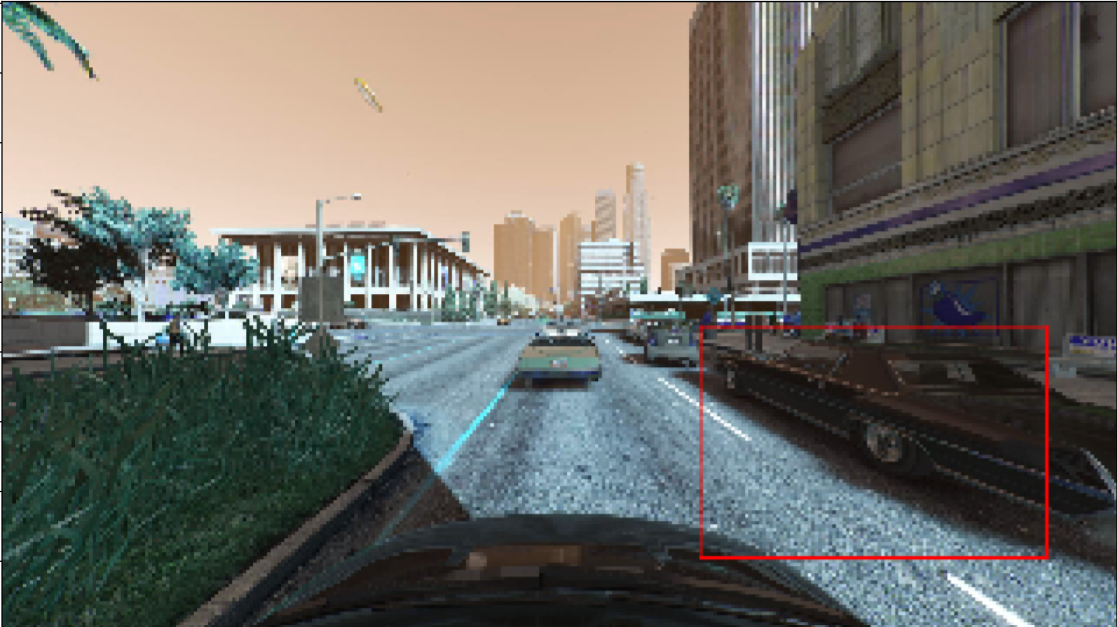}
\end{subfigure}
\begin{subfigure}[b]{.49\linewidth}
\includegraphics[width=\linewidth]{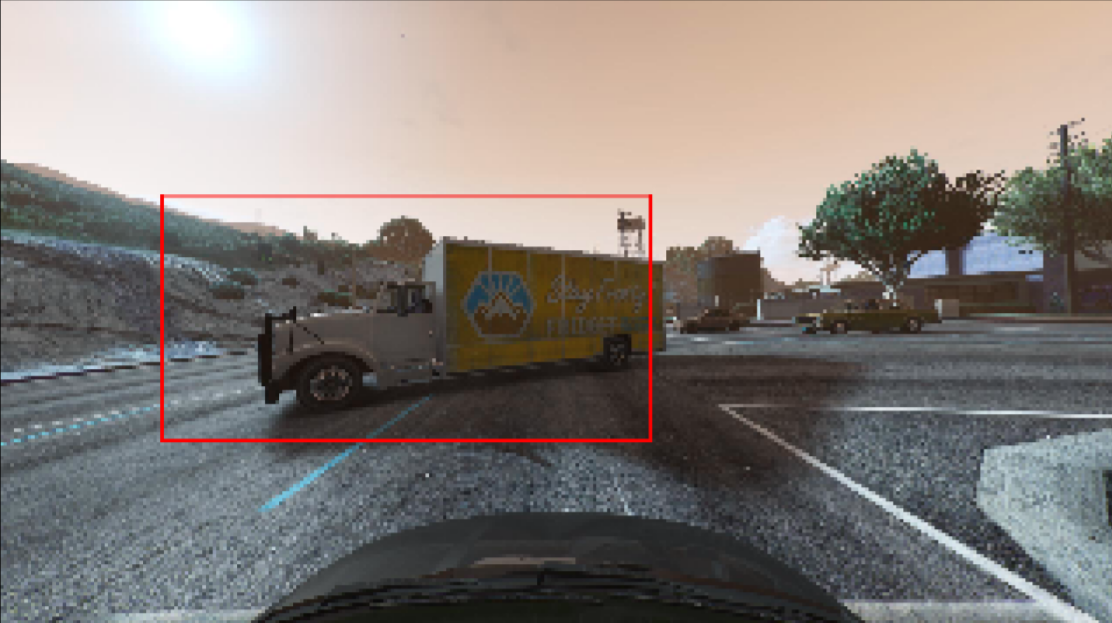}
\end{subfigure}

\begin{subfigure}[b]{.49\linewidth}
\includegraphics[width=\linewidth]{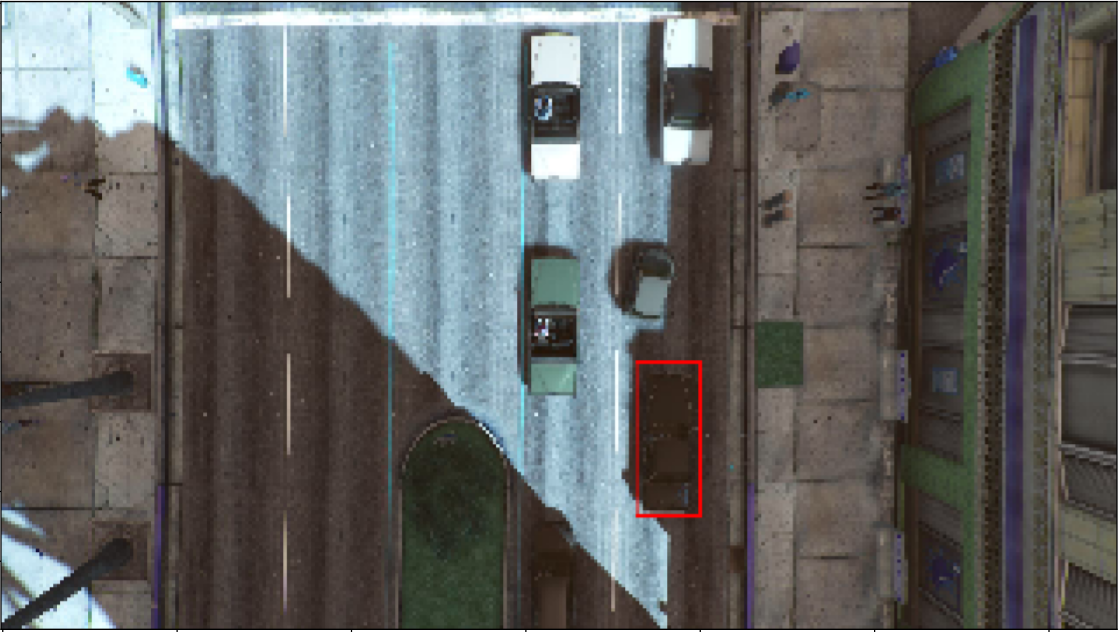}
\end{subfigure}
\begin{subfigure}[b]{.49\linewidth}
\includegraphics[width=\linewidth]{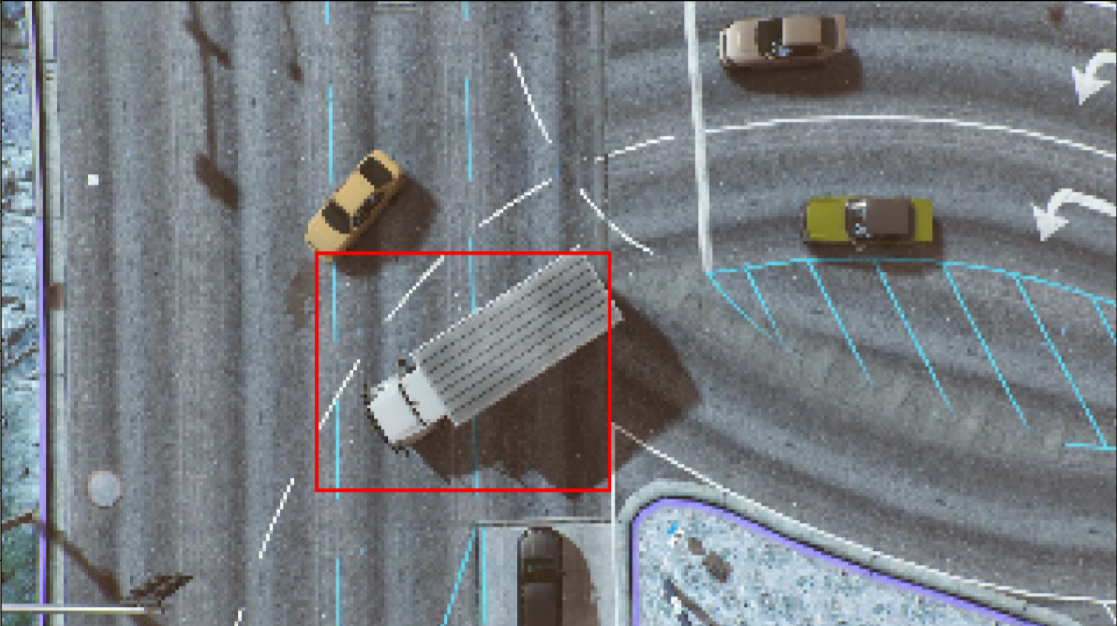}
\end{subfigure}

\caption{Sample images of the SAV dataset. Top images show the front-view, and bottom images are corresponding to their bird's-eye view~\cite{Palazzi2017}.}
\label{fig:SAV_Datase}
\end{figure}

The SAV dataset is a synthetic dataset created by Palazzi et al. \cite{ Palazzi2017 }. They have used a script to extract annotated images automatically from the Grand Theft Auto V (GTAV) video game. Images are taken from to frontal view and bird-eye view at each game time step. Sample images from the SVA dataset are illustrated in Figure \ref{fig:SAV_Datase}. There is an annotation file that contains spatial occupancy data for each vehicle, such as front and top view bounding boxes, depth, and rotation. Theses data have more than 1M annotated syntactic images.

\begin{figure}[th]
    \centering
    \includegraphics[width=1\columnwidth]{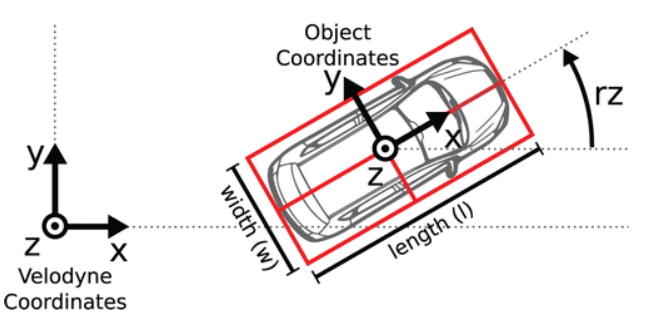}
    \caption{ The coordinate system of the annotated 3D bounding boxes in KITTI dataset. In the $z$ direction, the object coordinate system is located at the bottom of the object (contact point with the supporting surface)~\cite{geiger2013vision}. Note that the parameter denoted by~$\mathsf{rz}$ in this figure refers to yaw parameter of the object, represented by $\mathsf{rot}_Y$ in Eq.~\eqref{eq:annot}.}
    \label{fig:kiiti_coord}
\end{figure}

\begin{figure}[h]
\centering

\begin{subfigure}[b]{.49\linewidth}
\includegraphics[width=\linewidth]{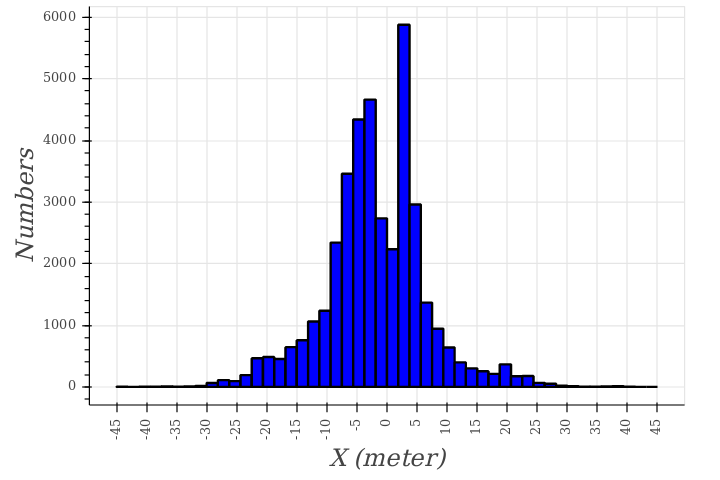}
\caption{$X$ Distribution }\label{fig:loc_x}
\end{subfigure}
\begin{subfigure}[b]{.49\linewidth}
\includegraphics[width=\linewidth]{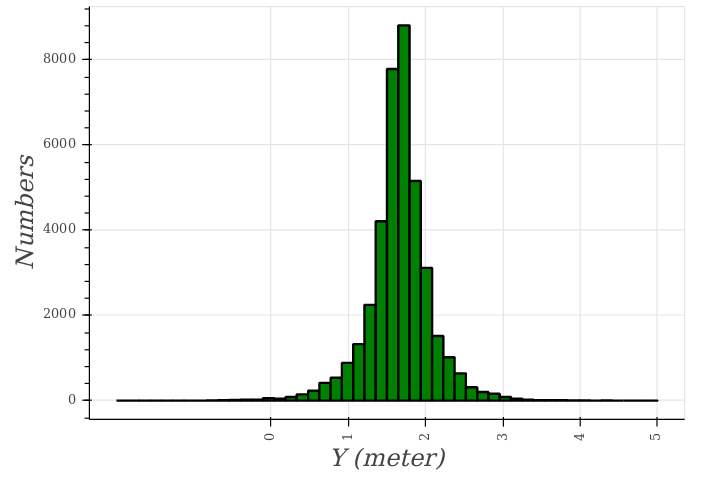}
\caption{$Y$ Distribution}\label{fig:loc_y}
\end{subfigure}

\begin{subfigure}[b]{.49\linewidth}
\includegraphics[width=\linewidth]{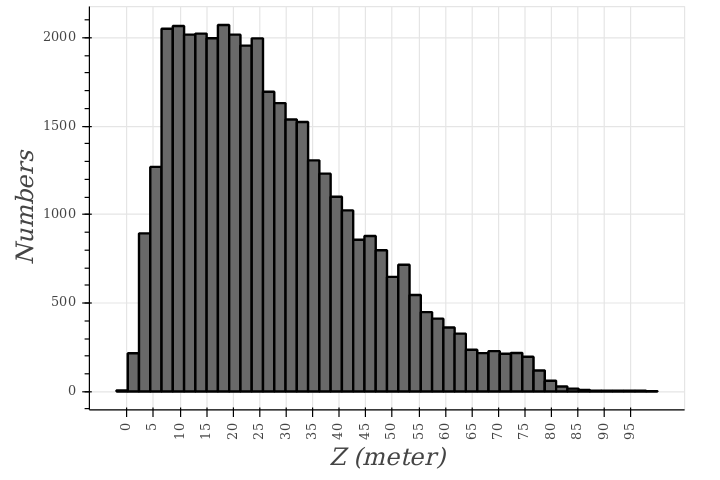}
\caption{$Z$ Distribution}\label{fig:loc_z}
\end{subfigure}
\begin{subfigure}[b]{.49\linewidth}
\includegraphics[width=\linewidth]{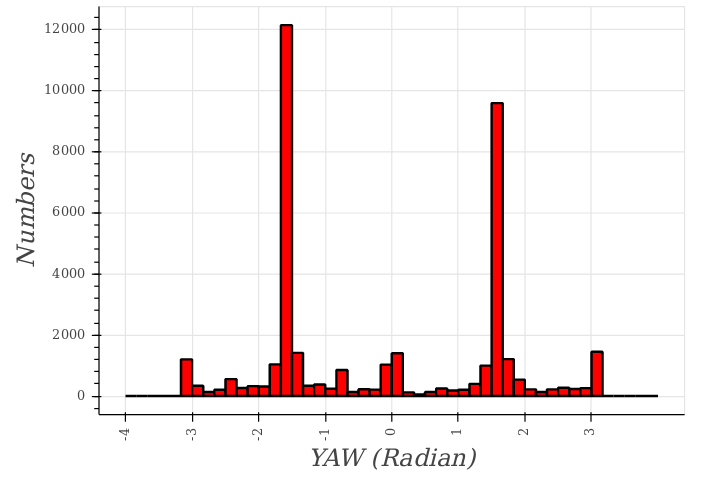}
\caption{$\mathsf{Yaw}$ Distribution}\label{fig:rot_y}
\end{subfigure}

\begin{subfigure}[b]{.65\linewidth}
\includegraphics[width=\linewidth]{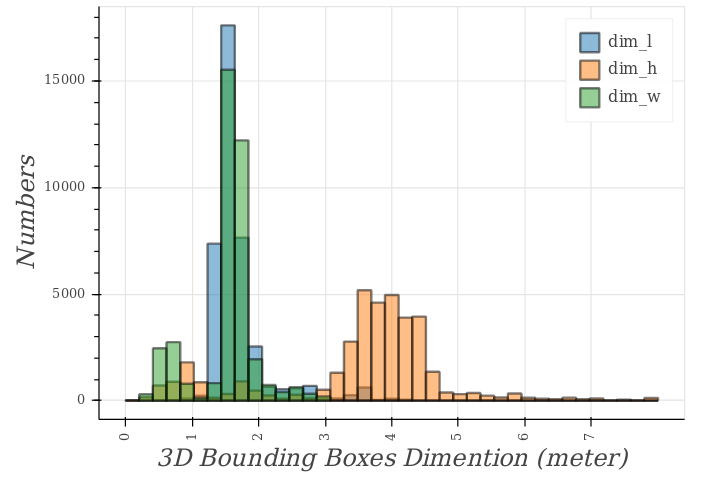}
\caption{Distribution of the size of the 3D bounding boxes (Width, Length, and Height)}\label{fig:dims}
\end{subfigure}

\caption{3D properties of KITTI dataset objects.}
\label{fig:kitty_static}
\end{figure}

Before going through the preprocessing procedure of images, we need to analyze the KITTI dataset to find how we can build appropriate targets for or regression network. We need to know the range of parameters and find outliers. We have shown a histogram of all parameters in Figure \ref{fig:kitty_static}. 
 the activation function of the last layer of the regression network is $\tanh (\cdot)$, whose range is in the interval $[-1,1]$.
Based on the range of this activation function, the first step is to map the annotations defined in Figure~\ref{fig:kiiti_coord} to the interval $[-1,1]$. In the prepossessing step, Eq.~\eqref{eq:annot} will be applied to annotations of the dataset:
\begin{subequations} \label{eq:annot}
\begin{equation}
X = \frac{\tilde{X}}{40}
\qquad \qquad
Y = \frac{\tilde{Y} - 2}{2} \\
\end{equation}
\begin{equation} 
Z = \frac{\tilde{Z}-50}{50}
\qquad \qquad
W = \frac{\tilde{W}-1.5}{1.5}
\end{equation}
\begin{equation} 
L = \frac{\tilde{L}-3.5}{3.5}
\qquad \qquad
H = \frac{\tilde{H}-1.5}{1.5}
\end{equation} 
\begin{equation}
 \mathsf{Yaw_{\sin}} = \sin(\mathsf{Yaw})
\end{equation} 
\begin{equation}
 \mathsf{Yaw_{\cos}} = \cos(\mathsf{Yaw})
\end{equation}
\end{subequations}
where $\tilde{p} = \left[ \tilde{X},\tilde{Y},\tilde{Z} \right]^T$ is the ground-truth center point location of the object in the annotations, $p = \left[ X,Y,Z \right]^T$ is the transformed center point location, $\tilde{s} = \left[ \tilde{W}, \tilde{L}, \tilde{H} \right]^T$ is the ground-truth dimensions of the object in the annotations, $s = \left[ W, L, H \right]^T$ is the transformed dimensions of the object, $\mathsf{Yaw}$ is the ground-truth yaw of the annotations, and $\mathsf{Yaw_{\sin}}$ and $\mathsf{Yaw_{\cos}}$ are the transformed yaw of the object.
This transformation allows us to estimate the properties of the 3D bounding boxes in the range of $[-1,1]$.


\subsection{Training}
Before training the network, data preprocessing is applied to the training set. The preprocessing procedure consists of the following steps. All detected vehicles are cropped from the front-view images based on their 2D bounding boxes and resized to $224 \times 224$ RGB images. The 2D bounding boxes are normalized based on the image sizes to be in the range $[-1,1]$. The same procedure will be applied to the 2D bounding boxes for top-view images. Random horizontal flip, brightness, contrast, hue saturation, and additive Gaussian noise are randomly applied to the training samples with probabilities of $0.25$~\cite{Buslaev2020Albumentations}. Data augmentation helps the network to avoid over-fitting and improves the model generalities by adding data diversity~\cite{shorten2019survey,Perez2017}. In this process, the drop-out layer with the probability of $0.25$ is applied after each fully-connected layer. Stochastically dropping-out neurons during training helps to avoid the co-adaptation of feature detectors~\cite {baldi2013understanding}.
As we have mentioned, our structure has four branches. The training procedure has two stages. In the first stage, the network is trained without the fourth branch to create a bird's-eye view map. In the next stage, all previously trained branches are frozen, and the fourth branch regressor is trained to estimate the properties of the object's 3D bounding box.

\begin{figure}[h]
\centering
\begin{subfigure}[c]{1\linewidth}
\includegraphics[width=\linewidth]{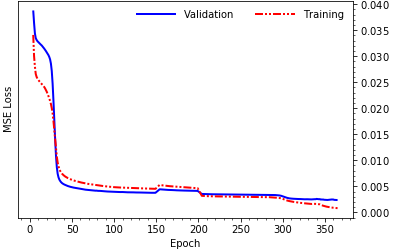}
\end{subfigure}

\begin{subfigure}[b]{1\linewidth}
\includegraphics[width=\linewidth]{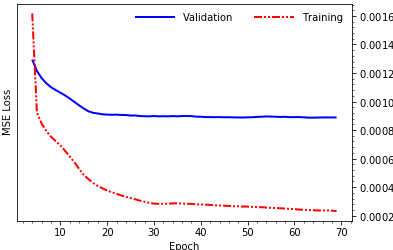}
\end{subfigure}
\caption{The top figure shows the MSE loss in the first stage of training the proposed network on the SAV dataset for 400 epochs. The bottom figure depicts the MSE loss in the finetuning of the proposed network on the KITTI dataset for another 100 epochs. Note that in the first stage of training, only the first three branches of the architecture shown in Figure~\ref{fig:Our-model} are
being trained.
}
\label{fig:MSE_error}
\end{figure}

At the first stage of training, we also freeze all layers of the pre-trained ResNet-50 network (i.e., the first branch, denoted by ``BR-1'' in Figure~\ref{fig:Our-model}), similar to the approach suggested in~\cite {Palazzi2017}. The second and third branches of the proposed network are trained using around half a million images in the SAV dataset for 400 epochs. The learning rate starts from 0.01 and is divided by ten every 100 epochs.
We consider mean square error (MSE) as a loss function, which is defined as 
\begin{equation}
\mathsf{MSE} = \frac{1}{n} \sum_{i=1}^{n} \left\| p_{\mathsf{2D}}^i - \tilde{p}_{\mathsf{2D}}^i \right\|^2,
\label{eq:mse_eq}
\end{equation}
where $n$ is the number of objects in each batch, and $p_{\mathsf{2D}}^i = \left[ x'_i, z'_i \right]^T$ and $\tilde{p}_{\mathsf{2D}}^i = \left[ \tilde{x'}_i, \tilde{z'}_i \right]^T$ are the estimated and ground truth values, respectively, of the corner points of the 2D bounding boxes of the $i$th object in the constructed bird's-eye view image.
We utilize stochastic gradient descent optimizer with momentum of $0.9$. Figure~\ref{fig:MSE_error} shows the loss variation during train the model in the stage-1.

\begin{figure*}
\centering 
\includegraphics[height=7cm]{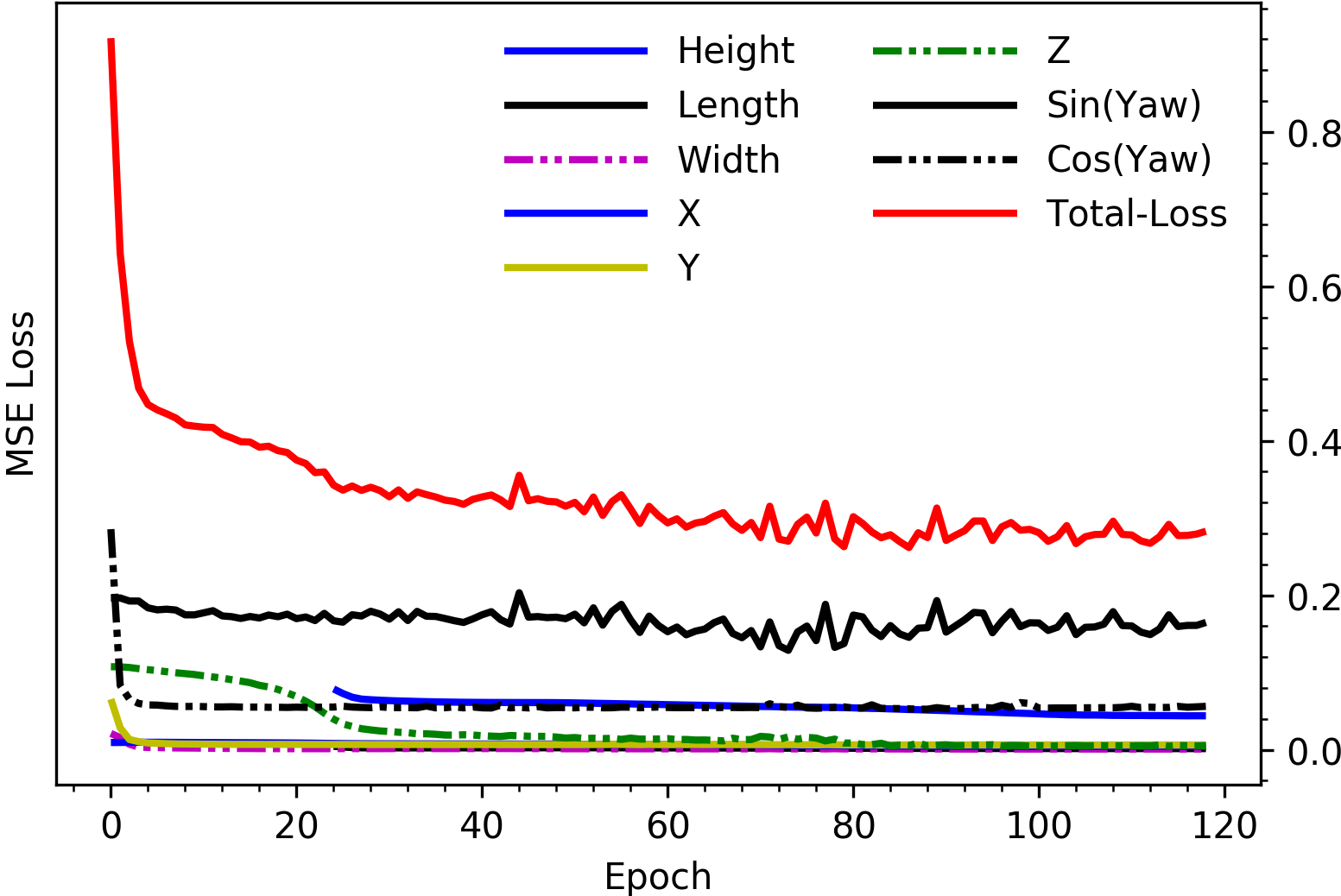}\par\medskip
\includegraphics[height=3.6cm]{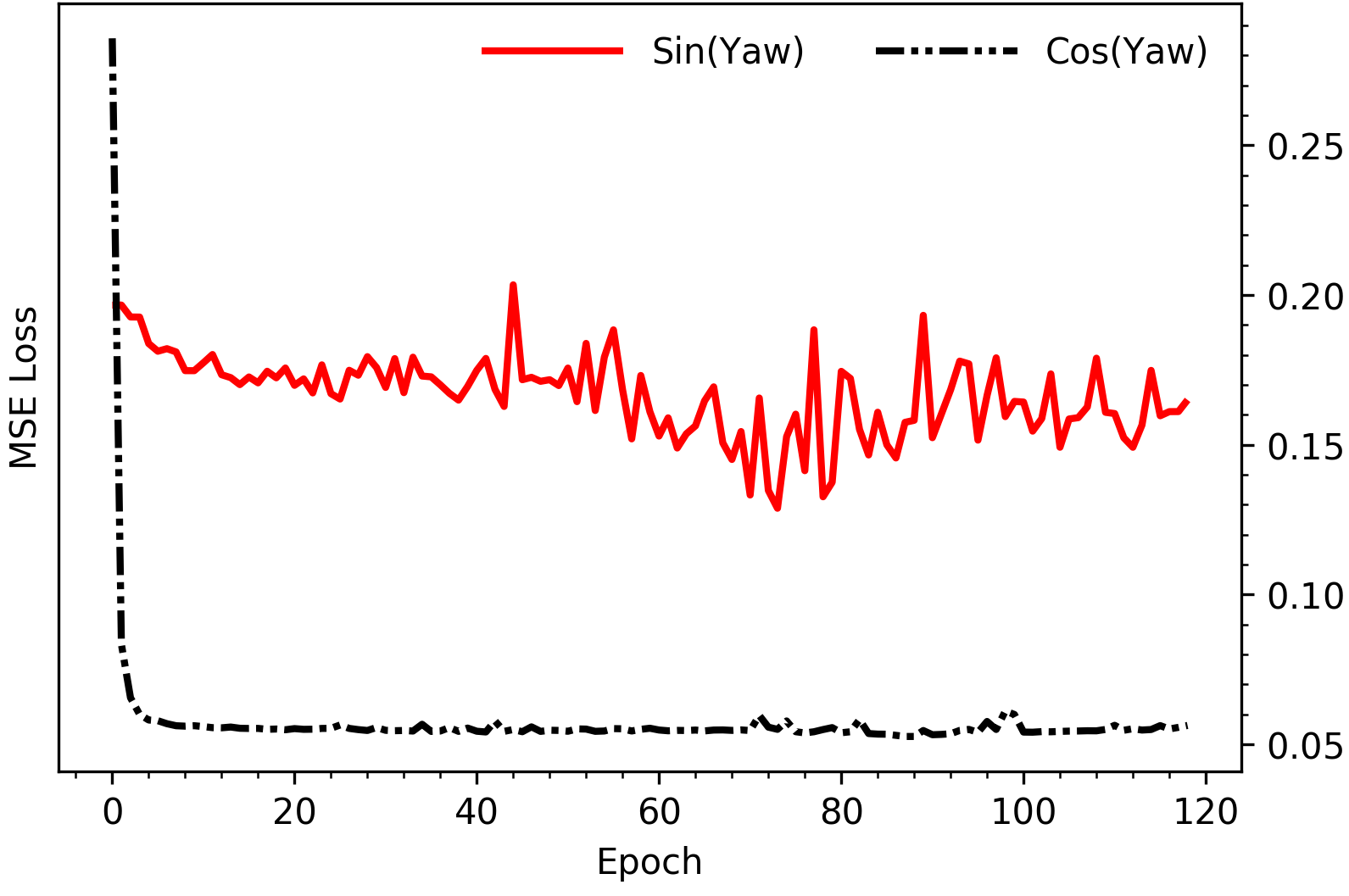}\quad
\includegraphics[height=3.6cm]{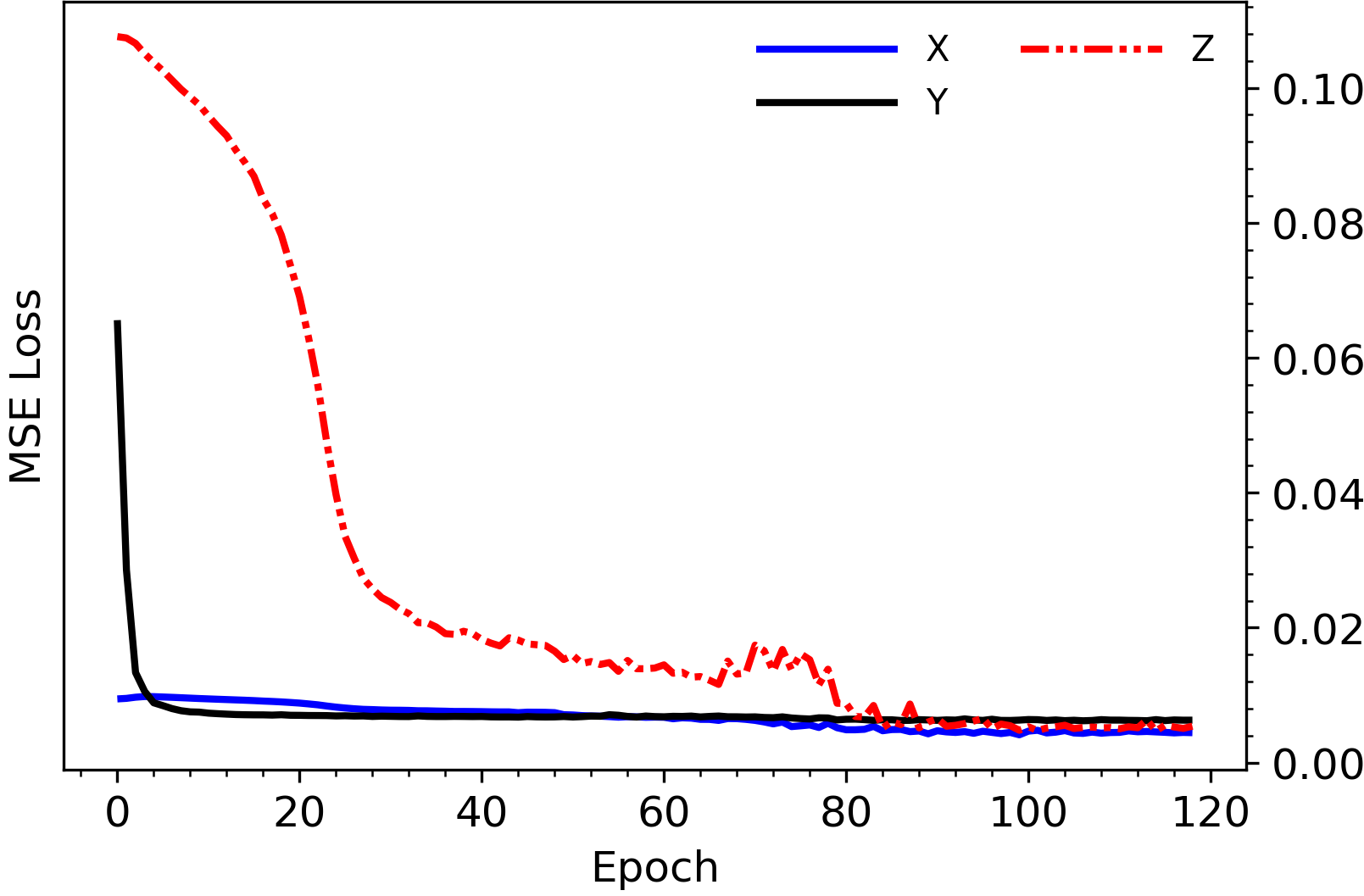}\quad
\includegraphics[height=3.6cm]{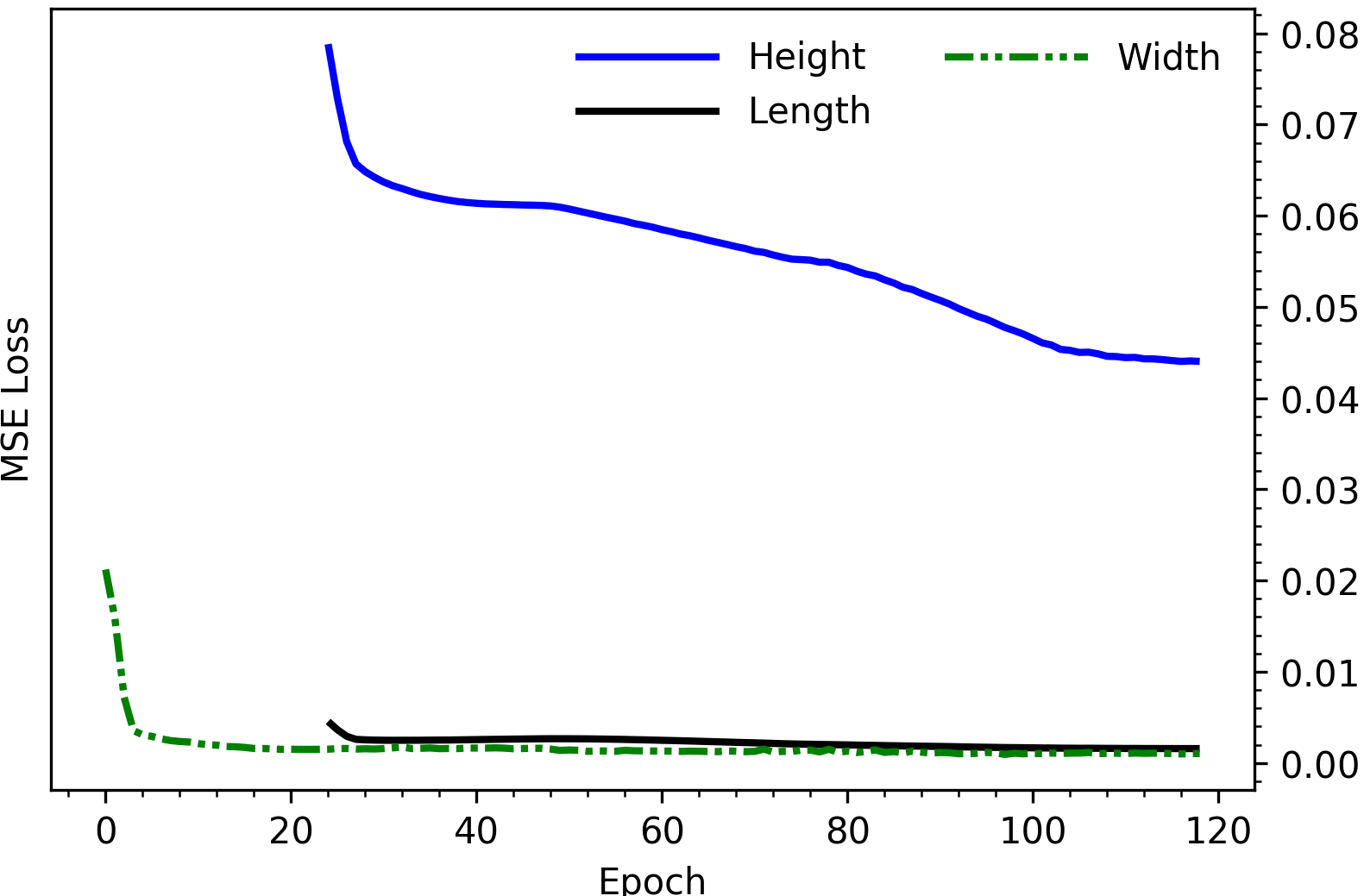}\par\medskip

\caption{Loss variations of the second stage of training, where the weights of the first three branches shown in Figure~\ref{fig:Our-model} are frozen and the last branch is trained on the KITTI dataset. The top plot shows losses of all estimated parameters and the cumulative loss versus the epoch number. The bottom figures show the loss of each parameter separately. From the left to the right, the MSE loss of yaw, 3D location of the center, and dimensions of the object are depicted.}
\label{fig:regress_losses}
\end{figure*}

In the second stage of training, the weights of the first three branches are frozen. The fourth branch is added, and the network is trained for 120 more epochs on the KITTI dataset to regress the 3D bounding box properties 3D of the objects. The loss variations in the second stage are shown in Figure~\ref{fig:regress_losses}. All networks presented in this paper are trained on a single GeForce Tesla V100 (32 GB) within three to four days. We use PyTorch V1.4.0 to implement the entire pipeline.

\subsection{Evaluation Metric}
The KITTI dataset contains three object classes, namely car, pedestrian, and cyclist. The dataset defines a standard for the level of difficulty of detecting each object based on the level of occlusion and truncation, and classifies objects in three levels of easy, moderate, and hard. Our proposed model is evaluated on the objects from all levels of detection difficulty. In our performance evaluations, we have used the training-validation split suggested in~\cite{chen20153d}, containing  about 3769 images, for both training and evaluation.
The standard KITTI dataset evaluation metric is used to assess the effectiveness of our proposed model.
More precisely, we have used the average precision~(AP) computed at three levels of intersection over union~(IoU) of $50\%$, $75\%$, and $90\%$ as an evaluation metric in our experiments to asses the proposed model. The AP is computed as a rough approximation of the area under the precision-recall curve~\cite{everingham2010pascal}, using 11-point sampling from the recall rates between $0.1$ and $1$.

\section{Results}
\label{Sec:Results}
\begin{figure}[ht]
\centering
\begin{subfigure}[b]{.9\linewidth}
\includegraphics[width=\linewidth]{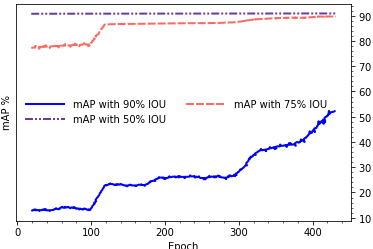}
\end{subfigure}

\begin{subfigure}[b]{.9\linewidth}
\includegraphics[width=\linewidth]{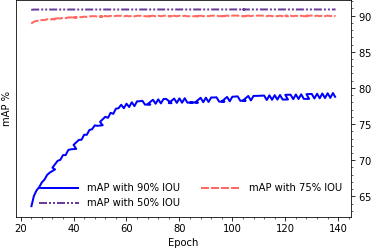}
\end{subfigure}
\caption{Evolution of mAP in the first stage of training the network for the estimation of the bird's-eye view maps. The top figure shows the evolution of the model while being trained on the SAV dataset for 400 epochs. The bottom figure depicts the evolution of the network when it is being fine-tuned on the KITTI dataset for additional 140 epochs.}
\label{fig:mAP_in_stage1_}
\end{figure}

To assess the performance of the first stage of training, the effectiveness of detecting the top-view 2D bounding boxes generated by BR-3 in the proposed network shown in Figure~\ref{fig:Our-model} is evaluated using mean AP (mAP), as depicted in Figure~\ref{fig:mAP_in_stage1_}.
As the dotted and dashed lines in Figure~\ref{fig:mAP_in_stage1_} show, around $90\%$ of the detected objects in the generated top-view images in the validation set have more than $50\%$ IoU with the ground truth.

\begin{figure*}
\centering 
\includegraphics[height=4cm]{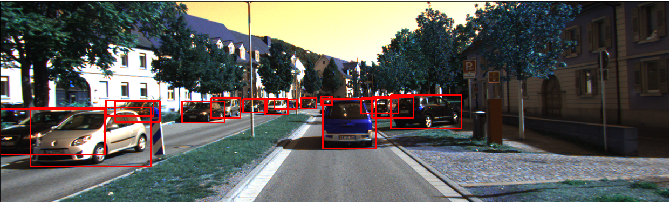}\quad
\includegraphics[height=4cm]{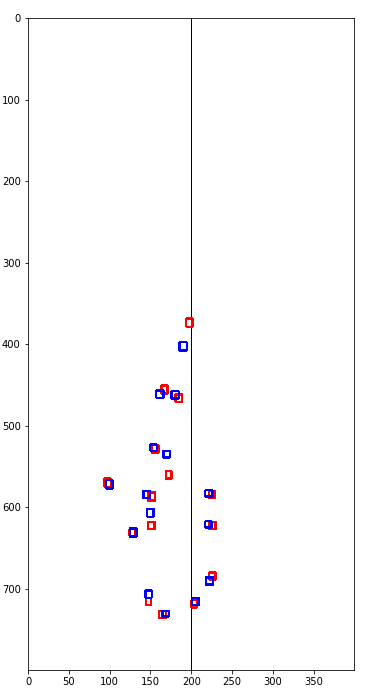}\par\medskip
\includegraphics[height=4cm]{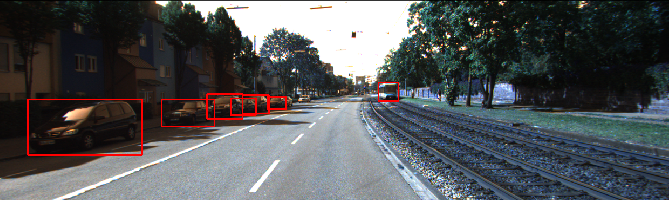}\quad
\includegraphics[height=4cm]{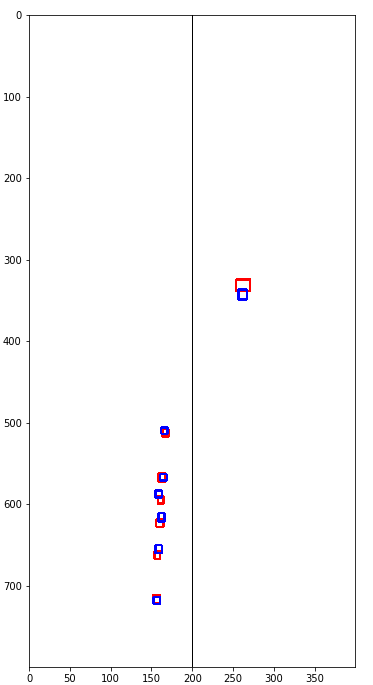}\par\medskip
\includegraphics[height=4cm]{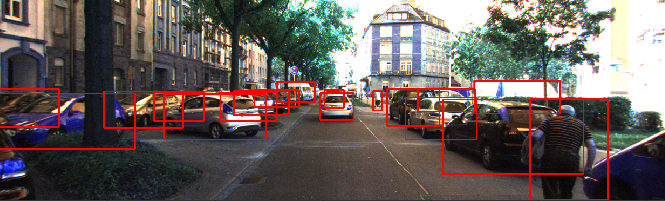}\quad
\includegraphics[height=4cm]{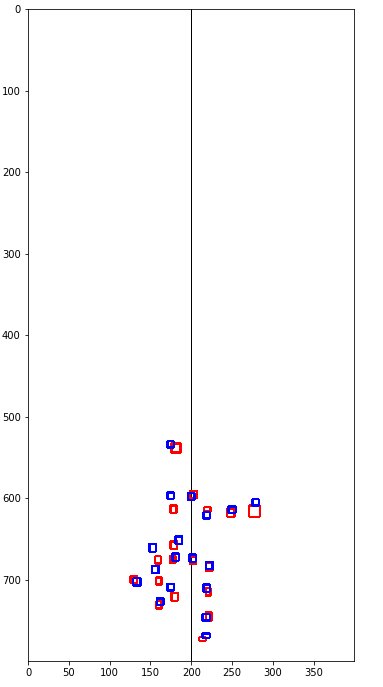}\par\medskip
\includegraphics[height=4cm]{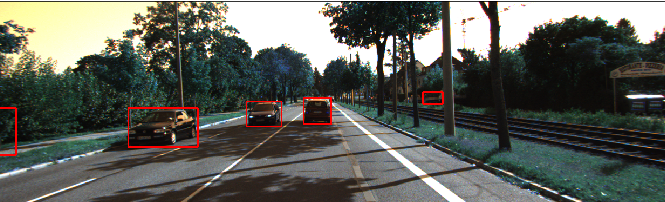}\quad
\includegraphics[height=4cm]{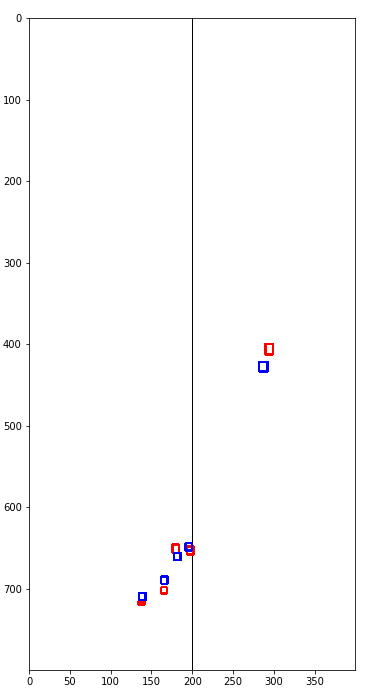}\par\medskip

\caption{Sample results for the images in the KITTI dataset. The images on the left show scenes with 2D bounding boxes, and the ones on the right illustrate the estimated bird's-eye view map of the same images produced by our proposed network. The blue rectangles are estimated top-view 2D bounding boxes, and the red rectangles show the ground truths.}
\label{fig:Fron_top_view_kitti}
\end{figure*}

An example bird's-eye view map with the top-view 2D bounding boxes is illustrated in Figure~\ref{fig:Fron_top_view_kitti}, where the front view images are shown on the left, and the corresponding top-view 2D bounding boxes associated with each object are depicted on the right.
A close observation of the top-view images on the right side of Figure~\ref{fig:Fron_top_view_kitti}, where blue and red rectangles show the estimated and ground-truth bounding boxes of the detected objects, respectively, shows that the majority of the instances have a high IoU.


\begin{figure*}
\centering 
\includegraphics[height=10cm]{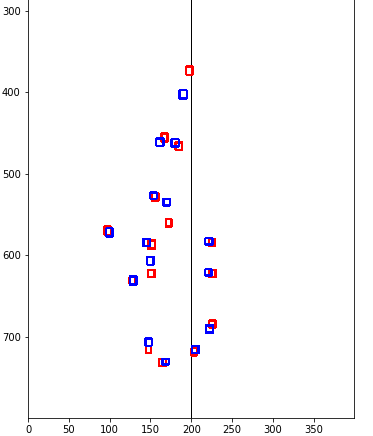}\quad
\includegraphics[height=10cm]{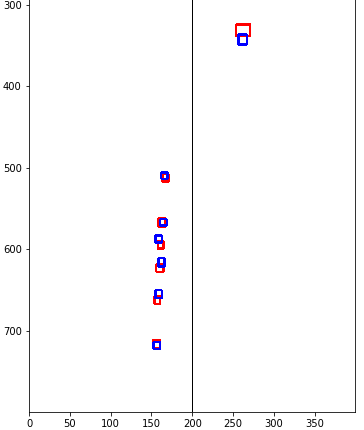}\par\medskip
\caption{The zoomed-in bird's-eye view images of the top two images shown in Figure~\ref{fig:Fron_top_view_kitti}. Blue  and red rectangles show the estimated and ground truth bounding boxes of the detected objects, respectively. As it can be seen, the majority of the instances have a high IoU.}
\label{fig:Top_view_kitti}
\end{figure*}




\section{Conclusions}
\label{Sec:Conclusions}
A network architecture was proposed in this paper to estimate the attributes of the 3D bounding boxes of objects detected in 2D images taken by monocular cameras. To that end, the network combines the constructed top-view bounding boxes of the detected objects with a deep representation of object features. The proposed model has a pre-trained ResNet-50 network as its back-end network and four more branches on top of it. The model first builds a top-view map of the scene to estimate the depth of the object and then estimates the object's 3D bounding boxes. We have implemented a two-stage training procedure to train this network on two major datasets: a syntactic dataset and the KIITI dataset.

\section*{Acknowledgment}
We express our appreciation to Dr.~Mohsin M. Jamali for his generous support and sharing his computational resources to make the process of training the CNNs discussed in this paper possible.

\bibliographystyle{IEEEtran}
\bibliography{refs}

\end{document}